\documentclass{article}

\usepackage[position, preprint]{neurips_2026} 

\usepackage[utf8]{inputenc}
\usepackage[T1]{fontenc}
\usepackage{hyperref}
\usepackage{url}
\usepackage{booktabs}
\usepackage{amsfonts}
\usepackage{nicefrac}
\usepackage{microtype}
\usepackage{xcolor}
\usepackage{graphicx}
\usepackage{amsmath} 
\usepackage{tcolorbox}

\title{Beyond Scaling: Agents Are Heading to the Edge}

\author{%
  Chunlin Tian\thanks{Equal contribution.}\\
  University of Cambridge\\
  University of Macau
  \And
  Dongqi Cai\footnotemark[1]\\
  Nanjing University
  \And
  Wanru Zhao\footnotemark[1]\\
  University of Cambridge
  \And
  Nicholas D. Lane\thanks{Corresponding author.}\\
  University of Cambridge
}

\begin{document}

\maketitle

\begin{abstract}
The bottleneck of useful agentic intelligence has shifted from compressing world knowledge into a single model to executing a coordinated system. 
This position paper argues that personal-agent architecture must move to the edge because the core properties of agentic intelligence tasks, particularly their structural coupling with high-fidelity local context and the need for zero-latency execution loops, do not sit well with cloud-centric designs. We develop this claim through three structural shifts. First, the Prefrontal Turn: the main marginal lever of capability has moved from pre-training scale to framework-level executive control. Such control must remain physically close to the environment of action if the agent is to preserve cognitive alignment. Second, the Data-Geography Paradox: the ``dark matter'' of agentic data (local file hierarchies, real-time sensor streams, and transient OS states) degrades, disappears, or loses meaning once prepared for cloud transmission, thereby cutting the agent off from ground-truth context. Third, the interaction-alignment loop: the only economically and ecologically sustainable source of agentic refinement data is the high-fidelity implicit preference signal produced through real-time local interaction. Third, the interaction-alignment loop: the only economically and ecologically sustainable source of agentic refinement data is the high-fidelity implicit preference signal produced through real-time local interaction. We conclude with falsifiable predictions for the next deployment cycle of personal agents.
\end{abstract}

\section{Introduction}
\label{sec:intro}
Scaling laws~\citep{kaplan2020scaling,hoffmann2022training} have served as the foundational blueprint for progress in large language models (LLMs). 
As model capabilities scaled predictably with parameters, data, and compute, scaling became the dominant route for pushing the Pareto frontier~\citep{henighan2020scaling,wei2022emergent}. However, as the field moves from conventional question-answering systems to interactive agents embedded in dynamic user environments, the limits of this ``knowledge-centric'' paradigm have become increasingly visible~\citep{ghaffary2024openai,morris2024position}. In agentic settings, long-term planning, tool orchestration, and environmental grounding become the primary constraints; these are not problems that can be solved simply by scaling up. The bottleneck of agency is therefore no longer the amount of compressed knowledge, but the structural alignment between the agent's executive control and the environment on which it acts.
We term this shift the ``Prefrontal Turn''~\citep{szczepanski2014insights, hazy2007towards}. Much like the functional dissociation in human cognition between posterior knowledge systems and the prefrontal executive apparatus~\citep{duncan1996intelligence,rascovsky2011sensitivity,goldman1995cellular}, the next leap in AI will not come from further compressing semantic knowledge into a single model. It instead requires a ``prefrontal'' substrate that can decompose tasks, manage memory, and self-correct in real time~\citep{sumers2023cognitive,shinn2023reflexion}. We argue that this turn requires architectural decoupling: the agentic framework must own the persistent task panels, skill libraries, and coordination logic.

This architectural evolution is driven by the Data-Geography Paradox: an agent's effectiveness is strictly bounded by its physical connection to the user's environment. Agentic workloads operate on what we call the `dark matter' of data: high-fidelity, volatile, and largely unindexed streams, including local operating-systems registries, real-time sensor data and fleeting interaction traces~\citep{shen2023hugginggpt,hong2023metagpt,xi2025rise,zeng2024agenttuning}. Such context cannot be archived or sent to the cloud in any meaningful form without substantial structural distortion and contextual loss. Moreover, the latency inherent in cloud-based retrieval-augmented generation (RAG)~\citep{lewis2020retrieval,barnett2024seven,zhou2023webarena} makes it difficult to preserve the spatio-temporal coherence required by agentic loops, often leading to interventions based on information that is already stale.

Beyond static execution, the edge is also the only ecologically valid site for the Interaction-Alignment Loop. We contend that agency is defined by real-time execution within volatile local environments, which is fundamentally incompatible with the latency and context loss imposed by remote architectures\citep{barnett2024seven,zhou2023webarena}. As the field moves toward decentralized multi-agent systems, the edge also provides a natural substrate for budget-aware swarms~\citep{moonshotai2026agentswarm,feng2024model}. In this paradigm, coordination reflects the distributed nature of modern devices, such as phones, PCs, and wearables~\citep{xu2024device,alizadeh2024llm}, each of which observes a different slice of local state.

Taken together, these claims define a broader architectural position. \textbf{This position paper argues that personal agents must become edge-native by structural necessity, rather than remain path-dependent artifacts of the cloud-centric scaling era.} We contend that the essence of agency lies in real-time execution within volatile local environments, which is fundamentally incompatible with the latency and context loss imposed by remote architectures. Consequently, the next leap in AI lies not in further scaling centralized knowledge repositories, but in anchoring decentralized executive control at the point of action.

\section{Agentic AI Background}
\label{sec:background}
\paragraph{Defining Agency through Environmental Coupling.}
We define an agent not simply as an LLM equipped with tool-use capabilities, but as an autonomous executive embedded in the user’s local digital and physical environment \citep{wang2024survey,wu2024copilot,wang2024mobile,rawles2023androidinthewild}. 
Unlike traditional chatbots that depend on static and global knowledge, personal agents derive their utility from perceiving and acting upon sovereign context, including private files, real-time sensor streams, and transient OS states \citep{morris2024position,ouyang2022training}. 
This definition deliberately excludes open-ended ``knowledge seekers'' and instead centers on ``environment actors'' that manage personalized daily workflows.
\paragraph{The Architectural Mismatch: Knowledge vs. Execution.}
Recent studies reveal a persistent ``agency gap'', in which frontier models with extensive declarative knowledge still struggle with basic long-term execution tasks \citep{valmeekam2023planning,kinniment2023evaluating,ghaffary2024openai}. 
We argue that this limitation arises not from insufficient reasoning ability, but from a structural mismatch between cloud-centric deployment and the low-latency demands of agentic tasks. 
Agentic systems require continuous, closed-loop interaction with their environment. 
However, existing architectures often treat the environment as a static state to be periodically sampled and transmitted \citep{ghaffary2024openai,lecun2022path}. 
This creates a ``grounding gap'', where the agent acts on stale snapshots of the user’s world.
\paragraph{The Blind Spots of Contemporary Benchmarks.}
Existing benchmarks, such as BFCL \citep{belcak2025small}, AgentBench \citep{liu2023agentbench}, and GAIA \citep{mialon2023gaia}, typically evaluate agents in idealized, static sandbox environments that miss the Data-Geography Paradox. 
These benchmarks fail to capture the structural degradation of high-fidelity context once it is serialized for remote inference.
They also ignore the ``coordination economy,'' namely the thermal, power, and memory overhead introduced by multi-step execution. 
As a result, strong benchmark performance does not necessarily translate into practical real-world utility. 
To address these limitations, agency should be redefined as a function of architectural proximity, shifting the focus from how much a model knows to how effectively it remains synchronized with its local environment.

\section{Edge Agentic: The Native Habitat}
\label{sec:edge}

\subsection{The Data-Geography Paradox: The Orbit Edge Case}
\label{sec:edge-paradox}

\begin{figure}[ht]
    \centering
    \includegraphics[width=0.85\linewidth]{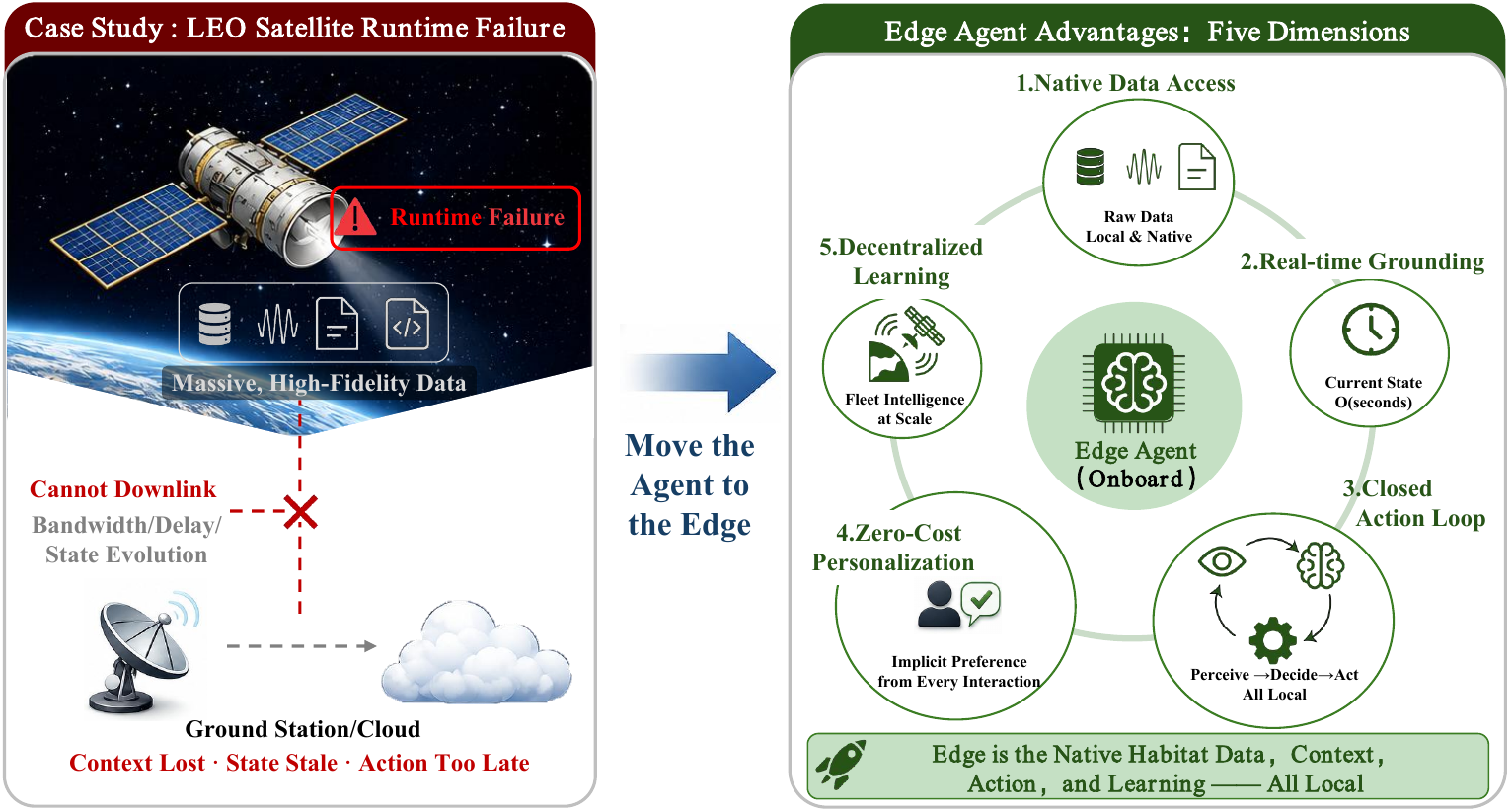}
    \caption{Edge Agentic vs. Cloud: LEO Satellite Failure.}
    \label{fig:case}
\end{figure}

To illustrate why agentic architectures must move toward the edge, as shown in Figure~\ref{fig:case}, we examine the structural limits of cloud-centric models in volatile environments. 
A representative case is runtime anomaly diagnosis in Low-Earth Orbit (LEO) satellites\citep{denby2020orbital,chien2005using}. 
Diagnosing OS-level memory leaks or attitude-control failures requires access to high-fidelity sensor streams and transient system logs, without aggressive compression or abstraction \citep{hundman2018detecting,xie2020satellite}. 
However, sending this context to the cloud introduces irreversible \textbf{spatio-temporal decoupling}. 
By the time telemetry is serialized, aggregated, and delivered to a remote model, the data has already undergone lossy abstraction, while the satellite may have already entered a different operational state.

This phenomenon illustrates the Data-Geography Paradox: agentic utility depends on architectural proximity to the environment in which actions occur. 
For an agent to remain ``situated,'' it must operate within the same temporal horizon as the state changes it seeks to manipulate. 
Cloud-centric architectures treat the environment as a static state to be periodically sampled, whereas edge systems perceive it as a continuous stream of evolving context \citep{shi2016edge,zhou2019edge,mcmahan2017communication}. 
Consequently, the primary bottleneck limiting cloud agents is not network bandwidth, but the structural decay of information that occurs when context is detached from its physical origin.

\subsection{What Edge Computing Uniquely Confers Upon Agents}
\label{sec:edge-advantages}
Edge computing is a \textit{structural imperative} for personal agency rather than an optional optimization. 
The five dimensions summarized in Table~\ref{tab:edge-advantages} address the core failure modes of remote inference across diverse environments, ranging from orbital computing systems to consumer devices \citep{xu2024device,alizadeh2024llm}. 
Together, they eliminate many of the persistent bottlenecks that constrain cloud-based agents.

\begin{table}[ht]
\caption{The five dimensions of edge-agent advantage. Edge-native architectures directly address the structural failure modes of cloud-based deployment.}
\label{tab:edge-advantages}
\centering
\small
\renewcommand{\arraystretch}{1.35}
\begin{tabular}{@{}p{0.25\linewidth} p{0.35\linewidth} p{0.35\linewidth}@{}}
\toprule
\textbf{Dimension} & \textbf{Cloud Failure Mode} & \textbf{Edge Advantage} \\
\midrule
\textbf{Native Data Access} 
& Context degradation through aggregated summaries 
& Direct access to raw and high-fidelity data streams \\

\textbf{Real-Time Grounding} 
& Temporal staleness caused by delayed retrieval and synchronization 
& Near real-time state awareness with $\mathcal{O}(\text{seconds})$ latency \\

\textbf{Closed Action Loop} 
& Network overhead and cumulative API costs 
& Atomic local execution with near-zero marginal cost \\

\textbf{Zero-Cost Personalization} 
& Dependence on aggregated and generic RLHF preferences 
& Implicit and in-situ personalization signals \\

\textbf{Decentralized Learning} 
& Reliance on homogeneous web-scale corpora 
& Diverse statistical learning across distributed environments \\
\bottomrule
\end{tabular}
\end{table}

\noindent \textbf{1. Native Data Access (Privacy \& Sovereignty).}
Personal agency depends on ``contextual dark matter''—high-fidelity sensor streams and unindexed OS states that cannot be serialized for cloud inference without substantial information loss. 
Because effective agents require deep access to sensitive user data \citep{carlini2021extracting,naveed2025comprehensive}, remote inference inherently conflicts with user trust and data-sovereignty regulations such as GDPR, PIPL, and DPDP \citep{regulation2016regulation,calzada2022citizens,naithani2025analysis}. 
Transitional architectures, including Apple’s PCC and Lenovo Qira \citep{apple2024pcc,nimbleedge2023,lenovo_qira}, partially reduce this exposure. 
However, the terminal architecture for personal agency must remain fully on-device. 
In this model, privacy becomes a structural property of the system rather than a policy guarantee.

\noindent \textbf{2. Real-Time Grounding.}
Agentic workloads are highly sensitive to state volatility in interactive environments \citep{xie2024osworld,liu2024mobilellm}. 
Dependence on remote infrastructure or high-latency Retrieval-Augmented Generation pipelines often produces functionally stale context, such as a meeting invitation modified only seconds earlier, leading to temporal hallucinations \citep{vu2024freshllms}. 
By executing locally, edge agents eliminate network round-trips and maintain $\mathcal{O}(\text{seconds})$ state latency. 
This enables interventions that remain grounded in the real-time state of the user’s environment.

\noindent \textbf{3. Closed Action Loop (Latency \& Cost Economics).}
Unlike single-turn chatbot interactions, autonomous tasks commonly require 50--100 sequential actions \citep{xie2024osworld,zhou2023webarena}. 
In cloud architectures, a persistent 200--500\,ms network overhead per step compounds into tens of seconds of perceptual dead time \citep{agrawal2023sarathi}. 
In contrast, local agents can achieve end-to-end voice-to-action latencies below three seconds, a responsiveness level difficult for remote systems to sustain \citep{liang2026comprehensive,chen2024octopus}. 
The economic constraints are equally significant. 
Maintaining dense and always-on execution loops through metered APIs rapidly becomes prohibitively expensive \citep{chen2023frugalgpt}. 
By treating local hardware as a sunk computational cost, edge systems drive marginal execution cost toward zero, making them the only practical substrate for high-frequency agentic workflows.

\noindent \textbf{4. Zero-Cost Personalization.}
Much of an agent’s practical capability emerges from post-training adaptation, and personalization at the individual-user level often yields utility far beyond that of a static global model \citep{fallah2020personalized}. 
Because autonomous workflows involve $N$ sequential reasoning steps, even small improvements in per-step accuracy compound across the execution trajectory. 
Conversely, globally averaged models systematically underperform on the long-tail distribution of any individual user’s workflows. 
Benchmarks such as LaMP \citep{salemi2024lamp} empirically demonstrate this personalization gap. 
Systems such as Hermes Agent \citep{nous2026hermes}, which employ closed learning loops through automatic skill generation and periodic consolidation, further illustrate the viability of continuous per-user adaptation. 
Delivering this capability through centralized cloud APIs remains difficult because maintaining per-user weight updates at scale introduces substantial memory overhead \citep{sheng2023slora}. 
On-device learning avoids this constraint. 
Every interaction, including retries, edits, and accepted outputs, naturally becomes an implicit preference signal acquired at near-zero marginal cost.

\noindent \textbf{5. Decentralized Learning \& Democratization.}
Although the interaction history of a single device is inherently limited, aggregating decentralized signals through privacy-preserving federated architectures \citep{zhuang2025} produces a large-scale statistical representation of diverse human behavior. 
This approach bypasses the emerging ``data wall'' and the increasingly homogeneous web-scraping pipeline that constrain centralized foundation models \citep{villalobos2022will}. 
At the same time, edge deployment democratizes the agentic ecosystem by reducing dependence on centralized infrastructure providers. 
Locally executing agents require neither permission from dominant model vendors nor reliance on fragile enterprise API contracts \citep{widder2023open}. 
As a result, computational and operational sovereignty shifts back to end users.

\subsection{Empirical Synthesis: Capability and Hardware Parity}

The transition toward edge-native agents does not require small language models (SLMs) to match the peak reasoning performance of frontier-scale models across all domains. 
Instead, we argue that SLMs have already crossed the operational sufficiency threshold for the dominant personal-agent workload distribution, including API orchestration, temporal scheduling, and structured summarization.
On structured tool-use and information-seeking benchmarks, models such as WideSeek-R1-4B \citep{xu2026wideseek} achieve performance comparable to the 671B-parameter DeepSeek-R1 \citep{guo2025deepseek}, despite a 170$\times$ parameter disparity. 
Similarly, compact models including Qwen3-4B, Phi-4-mini, and MiMo-7B \citep{xiaomi2025} remain highly competitive with frontier systems on many practical workloads. 
A substantial gap still persists in open-ended scientific reasoning, underspecified code synthesis, and tasks requiring deep world knowledge. 
However, our argument is narrower and more pragmatic: small models do not need to achieve universal parity with frontier models. 
They only need to reach functional parity within the task distribution that dominates personal-agent execution. 
At the same time, modern consumer silicon, including the Apple M4 Pro and Snapdragon 8 Elite \citep{qualcomm2026snapdragon,nvidia2026space}, already provides sufficient computational surplus to sustain local inference loops at approximately 30--80 tokens per second.


\subsection{The Deployment Gap: Why Agents Cannot Simply Be Ported to the Edge}
\label{sec:edge-gap}

Although the edge naturally provides these five advantages, existing cloud-native agentic frameworks cannot be directly transferred to satellites, smartphones, or other resource-constrained environments. 
This transition exposes four structural limitations in current agentic architectures.

\noindent \textbf{1. Violation of the Scale Assumption.}
Most cloud-native frameworks implicitly assume access to frontier-scale models with hundreds of billions of parameters. 
These systems tolerate loose prompting and inefficient planning because large models can often recover through brute-force reasoning. 
In contrast, edge environments, such as SWaP-constrained satellites, typically rely on SLMs. 
When an agent enters a trial-and-error hallucination loop, an SLM lacks the emergent capacity to recover reliably through scale alone.

\noindent \textbf{2. Coordination-Cost Blindness.}
Cloud architectures can aggressively fan out sub-agents because computation is amortized across large-scale data-center infrastructure. 
On edge devices, however, every \texttt{call\_subagent} invocation incurs an immediate cost in battery consumption, memory pressure, and thermal output. 
Current frameworks rarely model this ``coordination economy'' explicitly and therefore lack mechanisms for resource-aware orchestration.

\noindent \textbf{3. Memory-Architecture Mismatch.}
Cloud agents predominantly rely on remote vector databases and RAG pipelines for contextual access. 
Edge agents instead operate over heterogeneous and continuously evolving local data streams, including live telemetry, local file systems, and transient OS states. 
Under these conditions, memory cannot remain a remote retrieval layer. 
It must instead function as a deeply integrated local indexing and synchronization system.

\noindent \textbf{4. Reliability Through Brute Force.}
Cloud-based systems frequently depend on the broad world knowledge of frontier models as an implicit safety mechanism. 
For a 4B-scale model coordinating complex edge-side operations, this safety margin no longer exists. 
As a result, \emph{framework-level self-correction} becomes a hard requirement. 
An artificial anterior cingulate cortex-like mechanism must therefore be embedded directly into the agentic framework itself.

\vspace{-0.4em}
\begin{tcolorbox}[colback=green!5!white, colframe=green!40!black, boxrule=0.5pt, arc=0pt, outer arc=0pt, left=4pt, right=4pt, top=3pt, bottom=3pt]
\textbf{Key Points:} For an autonomous agent, the edge is not merely a ``faster'' or ``cheaper'' deployment option; it is its \textbf{native habitat}. The data, the temporal context, the execution loop, and the continuous learning signals all natively reside there. 
\end{tcolorbox}

\vspace{-0.8em}

\begin{tcolorbox}[colback=orange!5!white, colframe=orange!50!black, boxrule=0.5pt, arc=0pt, outer arc=0pt, left=4pt, right=4pt, top=3pt, bottom=3pt]
\textbf{Open Problems:} (1) Moving to the edge is an architectural transition, shifting from scaling posterior knowledge to engineering prefrontal executive control (the Prefrontal Turn, detailed in Section~\ref{sec:prefrontal}) (2) There is a critical lack of edge-native computational primitives, localized memory architectures, and framework-level self-correction (addressed in Section~\ref{sec:stack}).
\end{tcolorbox}
\vspace{-0.4em}

\section{Prefrontal Turn: Architecting the Executive Layer}
\label{sec:prefrontal}
\subsection{From Posterior Scaling to Prefrontal Orchestration}
We posit that the marginal lever for agentic capability has migrated from compressing world knowledge to engineering executive control. Over the past decade, scaling laws successfully built the artificial equivalent of the brain's \textit{posterior cortex}—a massive, monolithic substrate optimized for semantic memory, pattern recognition, and linguistic fluency. However, as agents move from passive QA to active environmental manipulation, raw posterior scaling is insufficient. 
The frontier now requires architecting the \textit{Artificial Prefrontal Cortex (PFC)}: a system dedicated to goal decomposition, working memory routing, and tool arbitration. We term this architectural shift the \textit{Prefrontal Turn}. It dictates that agency cannot spontaneously emerge from more pre-training compute; it must be explicitly engineered as a framework-level executive layer.

\subsection{Functional Dissociation: The Pathology of Monolithic LLMs}
To understand why scaling parameters fails to unlock these executive functions, a cognitive neuroscience analogy is instructive. Patients suffering from focal prefrontal damage exhibit a specific clinical dissociation: they retain fluent language and intact semantic memory, yet categorically fail at multi-step planning and goal execution. They retain the capacity to \textit{know}, but lose the capacity to \textit{do} 
Today's frontier cloud LLMs occupy precisely this pathological state. Because they are monolithic posterior systems without a structurally distinct executive apparatus, they are encyclopedically rich yet ``planning-fragile'' over compounding action sequences. The neurobiological lesson is that a functioning agent requires architectural separability. The prefrontal apparatus in AI demands its own computational primitives, independent memory indices, and, crucially, its own deployment locus. 

\subsection{The Edge as the Necessary Locus of Executive Control}
The \textit{Prefrontal Turn} necessitates an edge-centric paradigm due to the physics of control loops. Just as biological executive control requires closed-loop, near-zero-latency sensory feedback from the peripheral nervous system, artificial executive control requires synchronous access to the user's local state. 
Executive orchestration cannot be effectively ``bolted onto'' a remote cloud server. Operating the Prefrontal Cortex from a remote data center introduces structural failures: (1) Feedback Latency: Complex tool dispatch requires sub-second state verification, which is impossible under the cumulative network tax of cloud routing. (2) Contextual Entropy: The executive layer must maintain high-fidelity episodic memory of local OS states. Serializing this for cloud inference introduces irreversible information decay.
Thus, while the cloud may remain the repository of the artificial posterior, the edge is the inescapable locus of the prefrontal.

\subsection{The ``Artificial ACC'': Framework-Level Self-Correction}
Deploying the executive layer on SWaP-constrained edge devices introduces a new reliability challenge. Cloud models rely on massive parameter redundancy to absorb and ``brute-force'' reasoning errors. Edge models lack this luxury. 
Therefore, the most consequential research priority for the Prefrontal Turn is engineering the Artificial Anterior Cingulate Cortex (ACC)—a lightweight, framework-level module dedicated strictly to conflict monitoring, drift detection, and self-correction. To operationalize edge agents, this ``safety net'' must transition from being an emergent property of model weights to an explicit computational primitive within the edge-native agentic framework.

\vspace{-0.4em}
\begin{tcolorbox}[colback=orange!5!white, colframe=orange!50!black, boxrule=0.5pt, arc=0pt, outer arc=0pt, left=4pt, right=4pt, top=3pt, bottom=3pt]
\textbf{Resolution 1: The Structural Imperative of the Edge} \\
Edge migration is the physical manifestation of the Prefrontal Turn. Executive control must be physically anchored in the user's local environment to maintain a real-time grounding loop.
\end{tcolorbox}
\vspace{-0.4em}

\section{Framework Implications: Edge as the Native Substrate for Agent Swarms}
\label{sec:stack}
The Prefrontal Turn necessitates a structural reorganization of the agentic software stack. If executive control, persistent memory, and tool arbitration must reside near the environment, then existing agentic frameworks acquire a fundamentally different natural substrate. Edge deployment is not merely a viable hardware target for swarm-style agent systems; it is the physical environment in which their decentralized topologies achieve spatio-temporal coherence. This section therefore translates the position of Sections~\ref{sec:prefrontal} and~\ref{sec:edge} into a framework-level claim: \textbf{edge deployment is not merely compatible with swarm-style agent systems; it is the environment in which their decentralized structure becomes most meaningful.}

\subsection{Single edge agents as the baseline}

Before considering swarms, the edge already changes what a single agent can be. The emerging architectural pattern in agentic deployments is a three-layer separation: an \textbf{end user} interacts with a \textbf{task panel} that exposes schedules, triggers, status, permissions, and results; the \textbf{agentic framework} owns planning, memory, skills, tools, evaluation, sandboxing, and inter-agent coordination; underneath sit one or more \textbf{language models}, possibly small, multiple, and on-device.
\noindent \textbf{Users address task panels, not chat boxes.}
Recent agentic frameworks such as Hermes Agent~\citep{nous2026hermes} illustrate this separation. Users do not merely converse with a model; they trigger tasks across messaging clients, command-line interfaces, and workflow surfaces, then receive status updates and results. The framework, not the chat box, owns persistent memory, skill construction, sandboxed execution, tool invocation, and delegated sub-agents. In this architecture, conversation is one interface among many, not the defining system boundary.

\noindent \textbf{Layer decoupling makes the model swappable.}
Once the framework owns memory, skills, evaluation, and sandboxing, the model layer becomes a substitutable backend. A system can route among cloud APIs, local inference servers, and on-device SLMs without changing the surrounding agentic substrate. This decoupling is the architectural precondition for our edge claim. \emph{If the model layer is swappable, then choosing a SLM running on-device instead of a frontier LLM in the cloud becomes a deployment decision rather than a full-stack rewrite.} Recent on-device LLM work and surveys show that this baseline is technically plausible through small models, quantization, memory-aware inference, and mobile deployment frameworks~\citep{xu2024device,liu2024mobilellm,chen2024octopus,alizadeh2024llm}. For the modal personal workload, the single edge agent is therefore the immediate architectural implication: it keeps perception, action, and feedback within the user's local environment.

\subsection{From single edge agents to edge swarms}

Once the framework runs more than one agent, the immediate problem is coordination. Existing systems already explore hierarchical, graph-based, and swarm-style organization; our claim is not to invent these topologies, but to identify how their deployment assumptions change at the edge. Cloud-scale systems such as Kimi Agent Swarm show that parallel sub-agent orchestration can reduce latency and improve coverage for large, decomposable tasks~\citep{moonshotai2026agentswarm}. Separately, distributed inference systems such as Petals and LinguaLinked show that heterogeneous devices can participate in large-model inference through partitioning, fault tolerance, and load balancing~\citep{borzunov2023distributed,zhao2024lingualinked}. The open architectural opportunity is to combine these trends: not to copy a cloud swarm wholesale onto a phone, but to design smaller, budget-aware swarms whose agents are grounded in local devices and local context.

\noindent \textbf{Hierarchical (conductor / director-worker).}
A central orchestrator decomposes the task and dispatches sub-tasks to specialized worker agents, collects results, and produces a final output. WideSeek-R1~\citep{xu2026wideseek} is a representative instance: a lead agent issues \texttt{call\_subagent} invocations and merges the returns. Strengths include clear accountability, predictable resource use, and easy debugging. Weaknesses include bottlenecking at the orchestrator and a single point of failure.

\noindent \textbf{Swarm (decentralized).}
Agents communicate peer-to-peer with no single controller; coordination emerges from local interactions~\citep{moonshotai2026agentswarm,feng2024model}. This topology is often discussed as a general multi-agent framework, but it becomes especially natural at the edge for three reasons. First, edge environments are already distributed: phones, laptops, wearables, home servers, robots, and satellites each observe different local state. Second, the relevant actions are local and asynchronous, making centralized cloud arbitration a poor fit for many routine tasks. Third, the learning signal is decentralized by construction, since user corrections, retries, and acceptances are generated where the agent acts. Emerging work on distributed Mixture-of-Agents and LLM-enabled robot swarms makes this connection explicit: multiple LLMs can operate on individual edge devices and exchange information without a centralized server, while physical robot swarms can use local LLM instances for collaboration and human-swarm interaction~\citep{mitra2024distributedmoa,strobel2024llm2swarm}. The edge therefore gives swarm coordination a physical substrate rather than merely a software metaphor.

This does not remove the known failure modes of swarm systems. Responsibility overlap, uncontrolled message volume, and conflicting partial solutions remain real concerns. The implication is instead that edge-native swarms require explicit budgets for communication, energy, memory, and trust, together with framework-level conflict monitoring. In our terminology, the swarm needs an artificial ACC as much as it needs distributed workers.

\noindent \textbf{Hybrid (graph-based).}
A directed graph of agents with conditional edges, neither strictly centralized nor fully flat. The graph encodes who talks to whom and under what condition, providing both observability and parallelism. Work on language agents as optimizable graphs formalizes this view by representing agents, tools, and inter-agent communication as computational graphs whose edges can be optimized~\citep{zhuge2024language}. For near-term edge deployment, graph-constrained swarms may be the practical compromise: decentralized enough to exploit local context, but structured enough to bound coordination cost.

\noindent \textbf{A 2D taxonomy.}
We therefore classify multi-agent architectures along two orthogonal axes: control topology (hierarchical / swarm / graph) and deployment locus (cloud / edge), shown in Table~\ref{tab:taxonomy}. This taxonomy clarifies why edge agents are not merely cloud agents copied onto smaller hardware: the deployment locus changes the cost model and the data geometry of coordination itself.

\vspace{-0.3em}
\begin{table}[h]
\caption{A 2D taxonomy of multi-agent architectures by control topology $\times$ deployment locus. The edge-deployed cells remain under-explored; swarm and graph-based coordination are especially natural candidates because edge environments are physically distributed.}
\label{tab:taxonomy}
\centering
\footnotesize
\begin{tabular}{@{}p{0.21\linewidth}p{0.36\linewidth}p{0.36\linewidth}@{}}
\toprule
 & \textbf{Cloud-deployed} & \textbf{Edge-deployed} \\
\midrule
\textbf{Hierarchical} & Salesforce Agentforce, Microsoft Copilot & Useful for single-device task control \\
\textbf{Swarm}        & CAMEL, MetaGPT (research)                & Natural for distributed edge environments \\
\textbf{Graph (hybrid)} & LangGraph, AutoGen production            & Practical path to budget-aware edge swarms \\
\bottomrule
\end{tabular}
\end{table}
\vspace{-0.6em}

Two observations follow. \emph{First}, the cloud-deployed cells are better populated than the edge-deployed cells; this is a research opportunity rather than a closed design space. \emph{Second}, edge deployment changes the interpretation of swarm coordination. In a cloud data center, a swarm is mostly a software topology running over centralized infrastructure. At the edge, the agents, data sources, action spaces, and learning signals are already physically distributed. The topology therefore matches the substrate.

We therefore state a more precise framework implication: \textbf{edge-native agentic systems should be designed as budget-aware swarms or graph-constrained swarms, rather than as monolithic cloud agents merely compressed onto local hardware.} Hierarchical control remains useful inside a single device or high-stakes task loop, but it should be understood as one coordination primitive within a broader edge swarm, not as the only natural organizational form.

\subsection{Research consequences}


\begin{enumerate}
\item \textbf{Memory and skills are framework-level concerns, not model-level concerns.} Hermes-style closed learning loops (auto-generated skills from execution traces, periodic memory consolidation) live in the agentic framework, not in the model weights. This is exactly where per-user post-training will eventually live too. Edge devices, which see a single user's full interaction history, are the natural locus for this learning.
\item \textbf{Coordination protocols are the next standard to emerge.} The web standardized HTTP; agents have not yet standardized the equivalent. Open standards for agent-to-agent communication (extending early efforts like agentskills.io, MCP, Anthropic's agent protocol drafts) will determine whether edge agents can interoperate or remain siloed.
\item \textbf{Reliability becomes the binding constraint at small scale.} A 4B model running ten sub-agents must self-correct; it cannot brute-force the right answer with more parameters. The ACC analogue from Section~\ref{sec:prefrontal} (conflict monitoring and error detection) becomes the most consequential research area for the Prefrontal Turn to actually deliver.
\end{enumerate}

\vspace{-0.4em}
\begin{tcolorbox}[colback=orange!5!white, colframe=orange!50!black, boxrule=0.5pt, arc=0pt, outer arc=0pt, left=4pt, right=4pt, top=3pt, bottom=3pt]
\textbf{Resolution 2: Edge-Native Framework Primitives} \\
Addressing the lack of edge infrastructure requires abandoning cloud-assumed infinite resources for budget-aware orchestration. 
\end{tcolorbox}
\vspace{-0.4em}

\section{Counterarguments and Limitations}
\label{sec:counter}
\subsection{Counterarguments}
\noindent \textbf{Objection 1: ``Frontier tasks will always require frontier-scale cloud models.''}
This is correct, but it does not contradict our position. We argue for a \textit{Spatial Division of Intelligence}. The cloud remains the optimal substrate for asynchronous, posterior knowledge synthesis. However, the \emph{modal} personal-agent workload is an execution-heavy, knowledge-light problem. For these tasks, the bottleneck is not the depth of reasoning, but the speed and fidelity of the perception-action loop.

\noindent \textbf{Objection 2: Steelman: ``Hybrid, not edge.''}
The strongest opposing position is \emph{hybrid-first}: route context-bound steps locally, and reasoning-heavy steps to the cloud (e.g., Apple PCC~\citep{apple2024pcc}, Copilot+ PCs). We narrow our claim against this: hybrid is correct for the \emph{heavy tail} (deep research, novel coding), but edge-native is imperative for the \emph{modal} personal-agent invocation. As on-device models achieve parity on tool-use (Section~\ref{sec:edge}), this median shrinks strictly toward local execution. 
Furthermore, \emph{split-inference}---partitioning model layers between edge and cloud---fails to resolve the Data-Geography Paradox. Because agentic logic requires high-bandwidth local context (OS streams, unindexed files), this massive payload must still be prefilled on-device. Edge memory and KV-cache budgets are effectively exhausted before intermediate tensors can even be offloaded; the structural friction is merely displaced, not removed.

\noindent \textbf{Objection 3: Why hasn't this happened yet?}
A reasonable critic may observe: hardware readiness is documented and on-device models are interactive on shipping silicon (Section~\ref{sec:edge-advantages}), yet in 2026 the modal agentic invocation remains cloud-bound. We offer three honest reasons. First, the economic incentive is misaligned: model providers monetize per-token cloud inference, not on-device autonomy. Second, the distribution surface is owned by OS vendors: only Apple, Google, and Microsoft can credibly ship a system-level edge agent, and they move on multi-year cycles. Third, the framework-level executive layer (the prefrontal apparatus of Section~\ref{sec:prefrontal}) is genuinely not yet engineered to production reliability at small scale; the work in Sections~\ref{sec:stack} and~\ref{sec:agenda} is not rhetorical. The shift we predict is not blocked by missing technology; it is unfolding slowly because incentives, surfaces, and engineering must align in sequence.



\subsection{Limitations.}
Our argument has a clearly defined perimeter. The biological analogy of Section~\ref{sec:prefrontal} is functional, not anatomical, and load-bears only the dissociation between knowledge and executive control. The empirical evidence for SLM capability parity is concentrated in standardized benchmarks; long-tail production failures (cumulative state corruption, distribution shift across user environments) may not be captured. On data, we grant that synthetic data, multimodal corpora, and untapped non-English text extend the runway; we make the weaker claim that for \emph{agentic} alignment specifically---long, idiosyncratic, environment-grounded interaction traces---synthetic data is a poor substitute. Federated post-training on such traces remains an unsolved learning algorithm, marked as Research Agenda item~1 (Section~\ref{sec:agenda}); the position is that the data \emph{locus} is structurally inevitable, not that the algorithm is solved. Finally, our thesis privileges a distribution of agentic value spread across many small workflows; if value concentrates instead in a few frontier-scale tasks, cloud-first remains defensible for that subset, which we do not contest.

\section{A Research Agenda for Edge-Native Agency}
\label{sec:agenda}

To actualize the Prefrontal Turn and mature the edge-native agentic ecosystem, the research community must pivot from pure parameter-scaling to systems engineering. We identify six load-bearing challenges spanning learning, operating systems, hardware, and decentralized coordination.

\noindent\textbf{1. Continuous On-Device Alignment.} Shifting from centralized RLHF to continuous, per-user preference optimization at the edge requires algorithms that mitigate catastrophic forgetting during local adaptation. This must be coupled with privacy-preserving federated aggregation to pool idiosyncratic learning signals without exporting raw traces.

\noindent\textbf{2. The ``Artificial ACC'' under Compute Budgets.} Because SLMs lack the massive parameter redundancy used by cloud models to absorb reasoning errors, lightweight framework-level self-reflection and error-correction mechanisms remain a central open problem. Verification, rollback, conflict monitoring, and confidence estimation must fit within local compute limits.

\noindent\textbf{3. Agents as First-Class OS Primitives.} Edge agents should not remain isolated applications. A system-level agent requires new OS abstractions for granular capability sandboxing, autonomous lifecycle management, permission mediation, and hardware-secured persistent episodic memory.

\noindent\textbf{4. Hardware-Algorithm Co-Design for Agentic Loops.} Edge memory hierarchies must be co-optimized for multi-agent execution. Priorities include KV-cache multiplexing, local speculative decoding, and shared-memory routing when multiple specialized sub-agents time-share a single physical base model.

\noindent\textbf{5. Decentralized Interoperability Protocols.} The agentic web needs standardized semantic routing protocols for peer-to-peer capability discovery, secure credential delegation, task handoff, and result attestation across physically distributed edge devices.

\noindent\textbf{6. SWaP-Aware Evaluation.} Pure-accuracy leaderboards are insufficient for edge agents. Benchmarks should explicitly score latency, active energy draw, memory pressure, task throughput, and failure under intermittent connectivity, reflecting the Size, Weight, and Power constraints of the physical edge.

\section{Conclusion}
\label{sec:conclusion}
While the past decade of language modeling asked \emph{how much a model can know}, the next will be defined by \emph{how effectively a system can act}. This shift requires architecting the apparatus of agency, planning, coordination, reflection, and continuous learning, which we argue must natively reside at the edge for modal personal workloads. The edge uniquely resolves deployment bottlenecks by inverting cloud economics and guaranteeing structural proximity, while providing the exclusive data locus for zero-marginal-cost preference signals and real-time personal context. We therefore urge the community to redirect marginal investment away from frontier-scale pre-training toward targeted post-training, edge orchestration, and federated learning, recognizing that the future of everyday autonomous systems demands less model, and more mind.


\bibliography{refs}
\bibliographystyle{plain}




\end{document}